\documentclass{article}

\usepackage[preprint]{neurips_2026}

\usepackage[utf8]{inputenc} 
\usepackage[T1]{fontenc}    
\usepackage{hyperref}       
\usepackage{url}            
\usepackage{booktabs}       
\usepackage{nicefrac}       
\usepackage{microtype}      
\usepackage{xcolor}         
\usepackage{enumitem}       
\usepackage{amsmath}
\usepackage{cleveref}
\usepackage{multirow}
\usepackage{pifont}
\usepackage{graphicx}
\usepackage{amssymb}   
\usepackage{algorithm}
\usepackage{algorithmic}

\title{Correctness Is a Direction: Geometric Answer Selection in Language Models}

\author{
  Marcus Armstrong \And Navid Ayoobi \And Pradham Mummaleti \And Alexander Chulzhanov \And Arjun Mukherjee \\
  Department of Computer Science \\
  University of Houston \\
  Houston, TX 77204 \\
  \texttt{\{miarmstr, nyoobi, pmummale, achulzha\}@cougarnet.uh.edu, amukher6@central.uh.edu}
}

\begin{document}

\maketitle

\begin{abstract}
Answer correctness is encoded as a recoverable geometric direction in the
hidden states of language models.
We show that the mean displacement from incorrect to correct answer
representations, computed at approximately 70\% of model depth from fifty
labeled examples with no parameter updates, yields a scoring direction that
outperforms zero-shot log-probability scoring by up to +32.0 percentage
points on factual benchmarks (ARC-Challenge and MMLU) and by +38.1 to
+51.8 percentage points on TruthfulQA, across five models spanning 1B to
8B parameters in three architecture families (Llama, Qwen, Gemma).
The method requires one forward pass and one dot product per candidate; no
generation is performed at inference.
Applied as a hallucination detector on individual (question, answer) pairs,
the recovered direction achieves 0.693~AUROC versus 0.578 for log-probability
scoring.
We additionally find that correctness directions for factual reasoning, domain
knowledge, and calibrated truthfulness are near-orthogonal in representation
space, revealing that language models allocate geometrically independent
subspaces to qualitatively distinct notions of correct answer, with
architecture-dependent variation in the degree of separation.
This structure explains the observed transfer pattern---the direction
calibrated on factual questions transfers within task type but not across
it---and suggests that LLM calibration failures may reflect a routing
problem: the model's internal representation contains more correctness
signal than its output behaviour exploits.
\end{abstract}

\section{Introduction}
\label{sec:introduction}

Language models internally distinguish correct from incorrect answers, yet this
signal rarely influences their outputs.
When a model answers a factual question incorrectly, its hidden representation
at the layer encoding semantic content differs from the representation on a
correct answer in a consistent, recoverable direction.
The standard autoregressive generation pipeline ignores this distinction
entirely, selecting outputs token-by-token based on next-token probability
rather than any direct measure of semantic
correctness~\citep{brown2020language, ouyang2022training}.
The result is well-documented: large language models (LLMs) hallucinate with
high surface confidence, and users lack lightweight tools to identify
unreliable outputs without expensive secondary
inference~\citep{lin2022truthfulqa, ji2023survey}.

A growing body of work demonstrates that high-level semantic properties are
linearly encoded in LLM hidden
representations~\citep{alain2018understanding, belinkov2022probing,
park2023linear}.
\citet{zou2023representation} showed that directions corresponding to abstract
concepts---honesty, harm avoidance, emotional valence---can be extracted from
hidden states and used to steer model behavior without fine-tuning, suggesting
that diverse semantic properties leave consistent linear traces in the
representation space.
Whether \emph{answer correctness} specifically---the distinction between what
a model should say versus what it will say---is similarly encoded as a
recoverable geometric direction in single-model hidden states has not been
established.
We investigate this question directly.

We find that it is, and that the choice of extraction method matters
critically.
A \textit{contrastive} displacement---the mean difference between
correct-answer representations and wrong-answer representations, averaged over
just fifty question-answer triples---recovers a direction that dramatically
outperforms zero-shot generation as a scoring mechanism, with no parameter
updates and no generation at inference time.
The contrastive formulation proves essential.
The standard displacement (correct-answer representations versus
no-answer representations) yields high signal-to-noise ratio but low
discriminative accuracy, because it encodes \emph{answer-ness} rather than
\emph{correct-answer-ness}---a dissociation we characterize explicitly and
which prior work on semantic directions has not distinguished.
At inference, scoring a candidate answer requires a single forward pass and a
dot product.

Examining directions recovered across multiple task types reveals an unexpected
structural property: \textit{task-orthogonality}.
Directions calibrated on ARC-Challenge~\citep{clark2018think},
MMLU~\citep{hendrycks2021measuring}, and
TruthfulQA~\citep{lin2022truthfulqa} are near-orthogonal in both
Qwen-2.5-1.5B and Llama-3.2-1B (pairwise cosine similarities
$\lvert\cos\theta\rvert < 0.15$), with partial alignment in Gemma-2-2B.
Correctness is not a single geometric concept in LLM representations but a
structured set of independent, task-specific directions.
This accounts for an otherwise anomalous transfer pattern: a direction
calibrated on ARC transfers to same-domain MMLU subjects (+6 to +20~pp above
zero-shot) but fails to transfer to TruthfulQA, not because the method is
brittle, but because the model allocates factual correctness and calibrated
truthfulness to geometrically independent subspaces.

Evaluated across five models spanning 1B to 8B parameters in three
architecture families (Llama, Qwen, Gemma), with an apples-to-apples
zero-shot log-probability baseline computed on identical held-out test
examples, geometric scoring via our contrastive direction outperforms
zero-shot by up to \textbf{+32.0 percentage points} on factual benchmarks
and by \textbf{+38.1 to +51.8 percentage points} on TruthfulQA.
On TruthfulQA---where zero-shot baselines range from 28.7\% to 39.5\%---
task-specific geometric scoring achieves 72.1\%--81.2\% with fifty calibration
examples and zero gradient computation.
Applied as a hallucination detector in a binary classification setting, the
recovered direction achieves 0.693~AUROC versus 0.578~AUROC for log-probability
scoring, a substantial gain from a signal that costs one dot product per
candidate at inference.

\paragraph{Contributions.} We make the following contributions:

\begin{itemize}

    \item \textbf{Contrastive geometric direction extraction.}
    We introduce a contrastive formulation for recovering correctness directions
    from single frozen LLM hidden states and demonstrate it substantially
    outperforms the standard correct-vs.-no-answer displacement.
    We characterize the dissociation between signal-to-noise ratio and
    discriminative accuracy, showing that each direction encodes semantically
    distinct information (\S\ref{sec:method}).

    \item \textbf{Calibration efficiency and probe parity.}
    We show that direction estimation saturates at approximately fifty labeled
    examples across all tested architectures, and that a logistic regression
    probe trained on the identical calibration examples matches GRM accuracy
    within 2.1~pp across all model-task combinations---confirming that the
    mean direction captures the full discriminative information available from
    fifty examples without optimisation (\S\ref{sec:method}).

    \item \textbf{Layer universality.}
    We identify that the optimal extraction depth is approximately 65--70\% of
    model depth across five models with different sizes and layer counts,
    suggesting a consistent representational transition from syntactic to
    semantic encoding of answer quality (\S\ref{sec:method}).

    \item \textbf{Task-orthogonal correctness geometry.}
    We characterize the geometric structure of correctness across task types,
    demonstrating that factual, domain-specific, and calibration-oriented
    directions occupy near-orthogonal subspaces, and providing an empirical
    account of when task transfer succeeds or fails (\S\ref{sec:geometry}).

    \item \textbf{Hallucination detection without fine-tuning.}
    We demonstrate 0.693~AUROC on binary correct/incorrect classification of
    (question, answer) pairs---substantially above the log-probability baseline
    (0.578~AUROC)---requiring no training beyond the fifty-example calibration
    step (\S\ref{sec:hallucination}).

\end{itemize}

\section{Related Work}
\label{sec:related}

\paragraph{Linear representations and probing.}
A substantial literature establishes that semantic properties are linearly
encoded in the hidden states of neural language
models~\citep{alain2018understanding, belinkov2022probing}.
Probing classifiers---linear models trained on top of frozen representations
to predict some property of the input---have been used to recover syntactic
structure~\citep{hewitt2019structural}, factual
knowledge~\citep{meng2022locating}, and sentiment across a wide range of
architectures~\citep{belinkov2022probing}.
The linear representation hypothesis formalises this empirically: features
corresponding to semantic concepts are not merely recoverable by linear
methods but are themselves approximately linearly
structured~\citep{park2023linear, elhage2022superposition}.
Our work differs from the probing paradigm in a fundamental respect: we train
no classifier.
The correctness direction is computed as a mean displacement over fifty labeled
examples and applied via a single dot product, requiring no optimisation.

\paragraph{Directions in hidden states for truth and honesty.}
Several recent works extract directional signals from LLM representations
that correlate with truth or honesty without relying on external verifiers.
\citet{burns2023discovering} recover a latent truth direction without
supervision by exploiting the logical consistency of statement-negation pairs,
showing that models internally represent factual status even when they
generate the incorrect answer.
\citet{marks2023geometry} extend this, demonstrating that truth exhibits
emergent linear structure across model families and that the direction
generalises across distinct fact domains.
\citet{zou2023representation} extract directions for abstract concepts
including honesty and use them to steer model outputs at inference time;
\citet{li2024inference} similarly extract truthfulness directions from
attention heads and intervene on the forward pass to elicit calibrated
responses on TruthfulQA.
\citet{orgad2024llms} find that probing classifiers trained on hidden states
are skill-specific: probes transfer within similar task types (e.g.,\ factual
retrieval) but fail across qualitatively different tasks, leading them to
conclude that LLMs encode multiple distinct notions of truthfulness rather
than a universal mechanism.
Our work is closest to this line in motivation, but contributes three
advances not present in prior work.
First, we introduce a contrastive formulation that targets
\emph{correct-answer-ness} rather than abstract honesty or binary truth,
and show this is strictly necessary: the standard displacement encodes
answer-ness and is dramatically less discriminative---a distinction prior
work on semantic directions has not made.
Second, we provide a \emph{geometric} account of skill-specificity by
measuring cosine similarity between task-specific directions and showing
near-orthogonality, which explains why probes fail to transfer without
requiring trained classifiers; \citet{orgad2024llms} observe the failure
empirically but do not characterise its geometric cause.
Third, unlike steering approaches~\citep{zou2023representation,
li2024inference} that modify the forward pass during generation, we score
candidates without generation, and show the mean direction matches a trained
50-sample logistic regression within 2.1~pp---confirming the mean captures
the full discriminative signal available from fifty examples without
any optimisation.

\paragraph{Hallucination detection and confidence estimation.}
Detecting hallucinations in generated text has attracted considerable
attention~\citep{ji2023survey}.
Consistency-based methods generate multiple responses and measure
inter-sample agreement, either lexically~\citep{manakul2023selfcheckgpt} or
semantically~\citep{kuhn2023semantic}; both require many additional forward
passes.
\citet{kadavath2022language} prompt the model to evaluate its own outputs,
finding that self-assessment is partially calibrated but requires additional
generation.
Methods that use hidden states directly for hallucination detection have
emerged more recently~\citep{chen2024inside, chuang2023dola}, but typically
require training on hallucination-labelled data or specialised decoding
procedures.
Our hallucination detector requires neither: a direction calibrated on fifty
general question-answer triples produces 0.693~AUROC on binary
correct/incorrect classification, substantially above the log-probability
baseline, with one dot product per candidate.

\paragraph{Answer selection and scoring.}
For multiple-choice tasks, log-probability scoring---assigning each candidate
the log-probability the model assigns to its tokens given the
question~\citep{brown2020language}---is the standard zero-shot baseline and
forms our primary comparison throughout.
Self-consistency~\citep{wang2023self} improves factual accuracy by sampling
many chain-of-thought responses and taking the majority answer, but requires
tens of generation calls per question.
Minimum Bayes Risk decoding~\citep{eikema2020map} similarly selects the
candidate with highest similarity to the candidate pool, incurring heavy
generation costs.
Our method scores candidates from a single forward pass per candidate with no
generation, and outperforms log-probability scoring by up to +32.0~pp on
factual benchmarks and by +38.1 to +51.8~pp on TruthfulQA.

\section{Geometric Correctness Directions}
\label{sec:method}

We describe our method for extracting a correctness direction from a single
frozen language model and analyse two properties that bear directly on its
practical utility: the optimal layer at which to extract it, and the number
of calibration examples required.

\subsection{The Contrastive Correctness Direction}
\label{sec:contrastive}

\paragraph{Setup.}
Let $Q$ denote a question, $A^+$ its correct answer, and
$\mathcal{W} = \{w_1, \ldots, w_K\}$ the set of incorrect candidate answers.
For a frozen language model with $L$ transformer layers, let
$h_\ell(Q, A) \in \mathbb{R}^d$ denote the hidden state at layer $\ell$ and
the final token position of the concatenated prompt
\texttt{"Question: \{Q\}\textbackslash nAnswer: \{A\}"}.
We work with L2-normalised representations throughout:
$\tilde{h}_\ell(Q, A) = h_\ell(Q, A) \,/\, \|h_\ell(Q, A)\|$.

\paragraph{Standard and contrastive directions.}
Given a calibration set $\mathcal{D} = \{(Q_i, A^+_i, \mathcal{W}_i)\}_{i=1}^N$,
two natural displacement vectors can be computed at layer $\ell$:
\begin{align}
  \mathbf{d}_i^{\,\text{std}} &= \tilde{h}_\ell(Q_i,\, A^+_i)
                                  - \tilde{h}_\ell(Q_i,\, \varnothing)
  \label{eq:std} \\
  \mathbf{d}_i^{\,\text{con}} &= \tilde{h}_\ell(Q_i,\, A^+_i)
                                  - \frac{1}{|\mathcal{W}_i|}
                                    \sum_{w \in \mathcal{W}_i}
                                    \tilde{h}_\ell(Q_i,\, w)
  \label{eq:con}
\end{align}
where $\varnothing$ denotes an empty answer string.
The standard direction (\ref{eq:std}) measures the displacement when
\emph{any} answer is appended to a question.
The contrastive direction (\ref{eq:con}) measures the displacement
specifically from \emph{incorrect} answers to the correct one.
In both cases the direction estimate is
$\hat{\delta} = \bar{\mathbf{d}} \,/\, \|\bar{\mathbf{d}}\|$, where
$\bar{\mathbf{d}} = \frac{1}{N}\sum_i \mathbf{d}_i$.

We characterise the consistency of each estimate via signal-to-noise ratio:
\begin{equation}
  \mathrm{SNR} = \frac{\|\bar{\mathbf{d}}\|}{\sigma(\mathbf{d})}
  \label{eq:snr}
\end{equation}
where $\sigma(\mathbf{d})$ is the mean per-dimension standard deviation
across calibration examples.

\paragraph{Inference.}
Given a question $Q$ and candidate answers $\{A_1, \ldots, A_K\}$, each
candidate is scored by its projection onto the recovered direction:
\begin{equation}
  s_i = \hat{\delta}^{\top} \tilde{h}_\ell(Q,\, A_i), \qquad
  \hat{A} = \arg\max_i \; s_i
  \label{eq:score}
\end{equation}
No generation is performed at inference time.
Each candidate requires one forward pass and one dot product.

\paragraph{The SNR--accuracy dissociation.}
A counter-intuitive finding motivates the contrastive formulation: across
all tested models and layer depths, the standard direction consistently
achieves higher SNR than the contrastive direction (27--60$\times$
vs.\ 8--15$\times$), yet the contrastive direction substantially outperforms
in answer selection accuracy in every configuration tested.
Table~\ref{tab:std_vs_con} shows results at the validation-selected layer
for each sub-3B model.
The standard direction encodes \emph{answer-ness}---the generic geometric
change that occurs when any text is appended to a question---rather than
\emph{correct-answer-ness}.
High SNR in the standard direction reflects this consistency, but consistency
in the wrong direction is not discriminative.
The contrastive direction explicitly cancels the answer-ness component,
retaining only the displacement attributable to answer quality.

The exception is Llama-3.2-1B at layer fraction 0.75 (absolute layer 12 of
16), where standard and contrastive directions converge to cosine similarity
$\approx$0.99 and yield comparable accuracies (58.0\% and 53.2\%,
respectively).
We interpret this as a layer-specific phenomenon in which the Llama
architecture conflates answer-ness and correct-answer-ness at this depth;
we return to this in Section~\ref{sec:layer}.

\begin{table}[t]
\centering
\caption{Standard vs.\ contrastive direction at the layer selected via
held-out validation for each model.
Zero-shot log-probability on identical held-out test examples
(ARC-Challenge) provides the baseline.
Layer selection uses a 20\% validation split; all reported accuracies
are on the remaining 80\% test set.
Accuracy changes relative to zero-shot in parentheses.}
\label{tab:std_vs_con}
\vspace{4pt}
\small
\begin{tabular}{lcccccc}
\toprule
Model & Layer & Zero-shot & Standard $\hat{\delta}$ & Contrastive $\hat{\delta}$
      & Std SNR & Con SNR \\
\midrule
Llama-3.2-1B  & 0.60 & 35.0\% & 26.1\% {\small($-$8.9)}
                              & \textbf{44.7\%} {\small(+9.7)}
                              & 43.4$\times$ & 8.1$\times$ \\
Qwen-2.5-1.5B & 0.65 & 41.0\% & 35.1\% {\small($-$5.9)}
                              & \textbf{61.1\%} {\small(+20.1)}
                              & 49.4$\times$ & 9.0$\times$ \\
Gemma-2-2B    & 0.70 & 44.4\% & 27.6\% {\small($-$16.8)}
                              & \textbf{66.5\%} {\small(+22.1)}
                              & 27.2$\times$ & 10.1$\times$ \\
\bottomrule
\end{tabular}
\end{table}

\subsection{Layer Depth and Extraction Point}
\label{sec:layer}

\begin{figure*}[t]
  \centering
  \includegraphics[width=\textwidth]{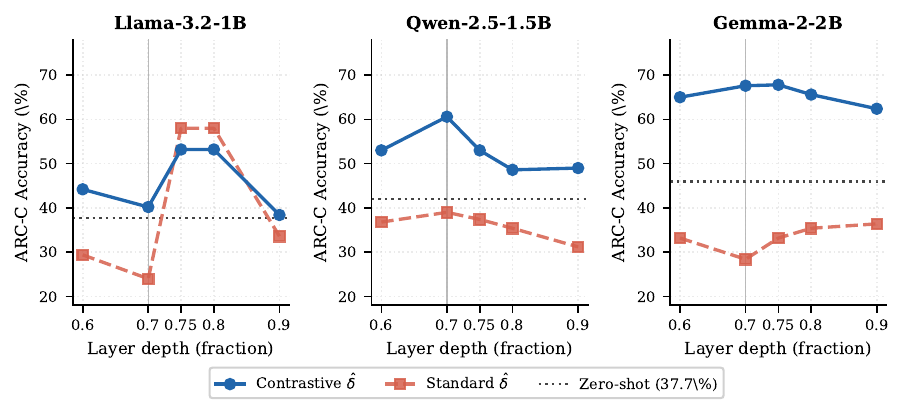}
  \caption{
    ARC-Challenge accuracy (\%) for standard $\hat{\delta}$ (orange, dashed)
    and contrastive $\hat{\delta}$ (blue, solid) across five layer fractions,
    for all three model families.
    The zero-shot log-probability baseline is shown as a dotted horizontal
    line for each model.
    The faint vertical guide marks fraction 0.70.
    Contrastive extraction outperforms standard at all tested depths for
    Qwen and Gemma.
    For Llama, the directions converge at fraction 0.75 (layer 12 of 16),
    where the standard direction reaches its global maximum of 58.0\%;
    at all other depths the contrastive direction is superior.
    Qwen and Gemma peak at fraction 0.70 despite differing architectures,
    consistent with a universal representational transition near this depth.
    Llama-3.2-1B points at fractions 0.75 and 0.80 are identical because
    both map to layer 12 in a 16-layer architecture.
  }
  \label{fig:layer_depth}
\end{figure*}

The representational content of LLM hidden states varies substantially across
depth.
Early layers process syntactic structure; later layers encode increasingly
abstract semantic content~\citep{tenney2019bert, belinkov2022probing}.
We evaluated contrastive direction extraction at seven fractional depths
(0.60, 0.65, 0.70, 0.75, 0.80, 0.85, 0.90) across all model families,
with layer selection performed on a 20\% held-out validation split to avoid
test-set contamination.
Results are shown in Figure~\ref{fig:layer_depth}.

\paragraph{Universal optimality at 65--70\% depth.}
Qwen-2.5-1.5B (28 layers) and Llama-3.1-8B (32 layers) peak at fraction
0.65; Gemma-2-2B (26 layers) and Qwen-2.5-7B (32 layers) peak at fraction
0.70; Llama-3.2-1B (16 layers) peaks at fraction 0.60 (absolute layer 9),
which corresponds to 56\% of depth but lies within one layer of the 65\%
point given the coarser granularity of a 16-layer model.
Across all five models the optimal extraction depth falls between absolute
layers corresponding to 56--70\% of total depth, despite architectures
ranging from 16 to 32 layers and 1B to 8B parameters.
We interpret this as evidence of a consistent representational transition
in this depth range where models shift from encoding syntactic and shallow
semantic features to encoding answer-quality-relevant geometry.

\paragraph{Standard direction underperforms zero-shot at shallow depth.}
At the selected layer, the standard direction underperforms zero-shot
log-probability for all three sub-3B models (Table~\ref{tab:std_vs_con}).
For Gemma-2-2B at fraction 0.70, the standard direction achieves only
27.6\% versus 44.4\% zero-shot---a $-16.8$~pp degradation.
This confirms that the standard direction encodes \emph{answer-ness} rather
than \emph{correct-answer-ness} even at the semantically richest extraction
depth: appending any text to a question moves hidden states in a direction
that is not only non-discriminative but actively anti-correlated with
correct-answer selection at these depths.
The contrastive direction at the same layer achieves 66.5\% (+22.1~pp above
zero-shot), confirming that subtracting the wrong-answer representations
isolates the quality-relevant component.

\paragraph{Llama convergence at fraction 0.75.}
As noted in Section~\ref{sec:contrastive}, at layer 12 of Llama, the
standard and contrastive directions converge to cosine similarity 0.99,
and both yield their highest accuracy for this model (58.0\% and 53.2\%).
The convergence suggests that at this specific depth the model has organised
its representations such that answer-ness and correct-answer-ness are nearly
identical---a property that holds at layer 12 but not at layer 11 (0.70) or
layer 13 (0.80, which maps to the same layer for Llama).
We note that this convergence is architecture-specific: it does not occur
for Qwen or Gemma at any tested depth.

\subsection{Calibration Efficiency}
\label{sec:calibration}

A practical concern for any calibration-based method is the number of
labeled examples required.
We compared $N{=}50$ and $N{=}100$ calibration examples across all three
models and both direction types (Table~\ref{tab:calibration}).
SNR is consistently higher at $N{=}50$ than $N{=}100$ for all models,
and downstream accuracy does not improve with doubling the calibration set.
We attribute this to two factors.
First, the mean direction estimate converges quickly because the per-example
deltas are highly consistent (SNR $>$7$\times$ even at $N{=}50$), leaving
little variance for additional examples to reduce.
Second, larger calibration sets drawn from an ARC train split with fixed
random seed increasingly include borderline questions whose displacement
vectors are noisy, marginally degrading the mean direction.

The saturation at $N{=}50$ has a practical implication: calibration requires
approximately 250 forward passes (50 examples $\times$ 5 passes each: one
for $Q{\oplus}A^+$, one for $Q{\oplus}\varnothing$, and three for
$Q{\oplus}w_j$) and completes in under five minutes on a consumer GPU for
the model sizes tested.
No gradient computation is involved.

\begin{table}[h]
\centering
\caption{Effect of calibration set size on direction SNR and ARC-Challenge
accuracy at the validation-selected optimal layer.
SNR and accuracy do not improve with doubled calibration set size.
Llama-3.2-1B results omitted from tabular comparison as the N=50 vs.\
N=100 systematic comparison was conducted at the previously identified
0.75 layer; the general pattern (no improvement with N=100) is consistent
across all models and layer configurations tested.}
\label{tab:calibration}
\vspace{4pt}
\small
\begin{tabular}{lccccc}
\toprule
& & \multicolumn{2}{c}{$N=50$} & \multicolumn{2}{c}{$N=100$} \\
\cmidrule(lr){3-4}\cmidrule(lr){5-6}
Model & Layer & SNR & Acc.\ (\%) & SNR & Acc.\ (\%) \\
\midrule
Qwen-2.5-1.5B & 0.70 &  8.4$\times$ & 60.6 &  8.4$\times$ & 60.6 \\
Gemma-2-2B    & 0.75 & 10.0$\times$ & 67.8 &  9.7$\times$ & 65.6 \\
\bottomrule
\end{tabular}
\end{table}

\section{Task-Specific Geometric Structure}
\label{sec:geometry}

The experiments in Section~\ref{sec:method} calibrate a direction on
ARC-Challenge and evaluate it on ARC-Challenge.
We now ask what happens when we move the calibration source, the evaluation
task, or both.
The answer reveals a structured geometric property of LLM representations that
we call \textit{task-orthogonality}: directions encoding correctness for
different task types occupy near-independent subspaces, with partial exceptions
that are themselves informative.

\subsection{Orthogonality of Correctness Directions}
\label{sec:orthogonality}

\begin{figure*}[t]
  \centering
  \includegraphics[width=\textwidth]{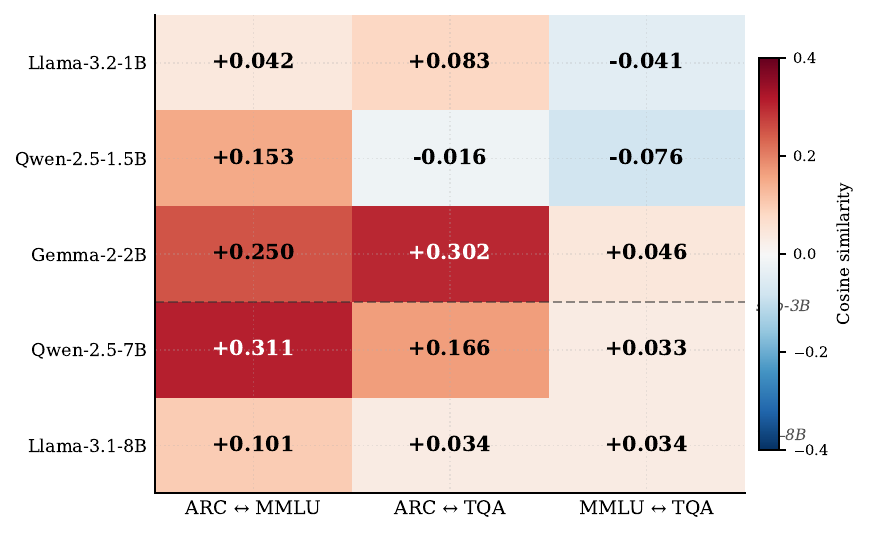}
  \caption{
    Pairwise cosine similarity between contrastive correctness directions
    calibrated on ARC-Challenge, MMLU-Biology, and TruthfulQA for all five
    tested models (rows), across the three task-direction pairs (columns).
    Warm colours indicate positive alignment; white indicates orthogonality;
    values are annotated directly in each cell.
    The dashed line separates sub-3B models (above) from 7B--8B models (below).
    Near-orthogonality ($|\cos\theta| < 0.16$) is the dominant pattern:
    Llama-3.2-1B, Qwen-2.5-1.5B, and Llama-3.1-8B satisfy this across all
    three pairs, with Llama-8B showing the strongest separation of any tested
    model (max $|\cos\theta| = 0.10$).
    Gemma-2-2B shows partial alignment between ARC and TQA ($+0.302$);
    Qwen-2.5-7B shows moderate ARC$\leftrightarrow$MMLU alignment ($+0.311$)
    and partial ARC$\leftrightarrow$TQA alignment ($+0.166$), indicating
    family-specific geometric organisation that can increase with scale.
  }
  \label{fig:cosine_matrix}
\end{figure*}

We compute contrastive correctness directions for three task types---factual
reasoning (ARC-Challenge), domain knowledge (MMLU, high-school biology
subject), and calibrated truthfulness (TruthfulQA)---and measure pairwise
cosine similarity between each pair.
Because all direction vectors are L2-normalised, cosine similarity equals
their dot product.
Results for all three model families are shown in
Figure~\ref{fig:cosine_matrix}.

\paragraph{Near-orthogonality as the dominant pattern.}
Across all five models, no task-direction pair exceeds cosine $+0.31$,
and most pairs satisfy $|\cos\theta| < 0.17$.
For the sub-3B Llama and Qwen models, all six off-diagonal similarities
satisfy $|\cos\theta| < 0.16$, indicating effectively independent subspaces.
This is not a trivial consequence of high dimensionality: if models used a
single generic correctness concept, all three directions would be strongly
aligned.
Instead, each task type has been allocated a geometrically distinct subspace.
At 7B scale, the pattern diverges by architecture family in an informative
way.
Llama-3.1-8B shows the strongest orthogonality of any tested model, with all
pairwise cosines at or below $0.10$---cleaner separation than its 1B
counterpart.
Qwen-2.5-7B moves in the opposite direction: ARC$\leftrightarrow$MMLU rises
to $+0.311$ and ARC$\leftrightarrow$TQA to $+0.166$, indicating that factual
and domain knowledge directions partially converge at this scale for the
Qwen family.

\paragraph{The Gemma exception.}
Gemma-2-2B shows substantially higher alignment: ARC $\leftrightarrow$ TQA
at $+0.302$ and ARC $\leftrightarrow$ MMLU at $+0.250$.
This partial alignment has a direct consequence for transfer: we show
in Section~\ref{sec:transfer} that the ARC direction yields $+12.9$~pp on
TruthfulQA for Gemma, versus $+4.8$~pp for Qwen and $+8.6$~pp for Llama.
The stronger alignment in Gemma suggests its representations organise
correctness at a coarser granularity than the other two models, potentially
a consequence of its larger parameter count (2B vs.\ 1--1.5B) or its
instruction-tuning procedure~\citep{team2024gemma}.
The Qwen-7B result---where ARC$\leftrightarrow$TQA rises from $-0.016$
(1.5B) to $+0.166$ (7B)---suggests that Gemma's partial alignment reflects
a family-specific property that can emerge with scale rather than an isolated
anomaly.
Both cases warrant investigation across a broader range of model families.

\subsection{Transfer Within and Across Task Types}
\label{sec:transfer}

The cosine similarity data predicts a specific transfer pattern: directions
should transfer between tasks whose geometric directions are aligned, and
fail---or invert---between tasks whose directions are orthogonal or
anti-correlated.
Table~\ref{tab:transfer} tests this prediction directly by evaluating an
ARC-calibrated direction on six target benchmarks spanning three semantic
categories.

\begin{table}[t]
\centering
\caption{
  Transfer of ARC-Challenge calibrated $\hat{\delta}$ to target tasks
  ($\Delta$~pp relative to zero-shot log-probability baseline on identical
  test examples).
  The bottom row shows TruthfulQA results when recalibrated on TruthfulQA
  examples (own $\hat{\delta}$), confirming that calibration source
  determines performance.
  Bold entries indicate positive transfer; negative entries indicate the ARC
  direction actively hurts relative to zero-shot.
}
\label{tab:transfer}
\vspace{4pt}
\small
\setlength{\tabcolsep}{5pt}
\begin{tabular}{llccc}
\toprule
Category & Task & Qwen-1.5B & Gemma-2B & Llama-1B \\
\midrule
\multirow{4}{*}{\textit{Factual MC}}
  & ARC-C (source)     & \textbf{+24.3} & \textbf{+20.3} & \textbf{+6.3}  \\
  & MMLU-Biology       & \textbf{+20.4} & \textbf{+19.7} & \textbf{+5.0}  \\
  & MMLU-Chemistry     & \textbf{+4.9}  & \textbf{+8.4}  & $-4.4$         \\
  & MMLU-World History & \textbf{+13.5} & \textbf{+22.4} & \textbf{+6.8}  \\
\midrule
\textit{Coherence}
  & HellaSwag          & $-11.0$        & $-16.7$        & $-26.7$        \\
\midrule
\multirow{2}{*}{\textit{Calibration}}
  & TruthfulQA (ARC $\hat{\delta}$) & +4.8 & \textbf{+12.9} & +8.6 \\
  & TruthfulQA (own $\hat{\delta}$) & \textbf{+41.5} & \textbf{+38.1} & \textbf{+44.9} \\
\bottomrule
\end{tabular}
\end{table}

\paragraph{Factual tasks: positive transfer.}
The ARC direction transfers positively to all factual multiple-choice tasks
for Qwen and Gemma.
MMLU-Biology receives $+19.7$--$+20.4$~pp above zero-shot from the
ARC direction, comparable to its own-calibration result ($+7.7$~pp for
Gemma, $+0.0$~pp for Qwen).
The ARC direction is a more effective calibration source than MMLU's own
validation split, which we attribute to calibration quality rather than task
mismatch: ARC's train split provides clean, unambiguous science questions
with clearly correct and clearly wrong distractors, whereas MMLU's small
validation splits (10--30 examples per subject) contain borderline examples
that add noise to the direction estimate.
World History, which is semantically distant from ARC's science focus, still
benefits substantially ($+13.5$--$+22.4$~pp), consistent with a direction that
captures general factual correctness rather than domain-specific knowledge.

\paragraph{Coherence tasks: negative transfer.}
HellaSwag, a sentence-completion benchmark requiring commonsense coherence
rather than factual retrieval, shows strongly negative transfer for all
three models ($-11.0$ to $-26.7$~pp).
The ARC direction actively degrades performance below zero-shot.
When calibrated on HellaSwag directly, the own direction achieves $+6.3$~pp
(Qwen) and $-4.7$~pp (Gemma) relative to zero-shot, confirming that
HellaSwag accuracy is not well-served by geometric scoring in general at
this calibration size.
We interpret this as evidence that the geometric signal for commonsense
coherence is orthogonal---or slightly anti-correlated---to the geometric
signal for factual correctness.
Completing a sentence naturally requires fluency and world-model consistency,
which may be encoded in different representational subspaces than declarative
fact retrieval.

\paragraph{Calibration tasks: near-zero ARC transfer, large own-direction gains.}
TruthfulQA tests whether models produce calibrated, honest answers rather
than confidently incorrect ones.
The ARC direction yields modest gains under argmax scoring ($+4.8$--$+12.9$~pp),
but the TruthfulQA-calibrated direction under argmax scoring yields
$+38.1$--$+51.8$~pp---the largest single-task improvement in our
experiments.
This confirms that TruthfulQA has a distinct correctness direction that is
only weakly aligned with the ARC direction in Qwen and Llama (cosine
$|{<}0.09|$) and partially aligned in Gemma (cosine $+0.302$).
Importantly, the TruthfulQA direction does not require argmin scoring: the
contrastive formulation, computed on TruthfulQA (correct, wrong) pairs,
extracts a direction that already points toward calibrated answers rather
than overconfident ones.

\subsection{Multi-Dimensional Correctness}
\label{sec:taxonomy}

The transfer results are parsimoniously explained by a simple model: LLMs
allocate geometrically independent subspaces to qualitatively distinct
notions of answer quality.
We identify three such dimensions in the tasks studied:

\begin{itemize}
  \item \textbf{Factual correctness} --- whether a statement corresponds to
        a retrievable fact.
        Encoded in the ARC and MMLU directions; these are moderately
        correlated (cosine $+0.04$--$+0.31$) and transfer positively
        within the category.
  \item \textbf{Commonsense coherence} --- whether a sentence completion is
        plausible and grammatically natural.
        Encoded in a direction approximately orthogonal or anti-correlated
        to factual correctness, as evidenced by HellaSwag's negative
        transfer.
  \item \textbf{Calibrated truthfulness} --- whether a model's answer
        reflects appropriate epistemic confidence.
        Encoded in the TruthfulQA direction, near-orthogonal to factual
        correctness for Qwen and Llama.
\end{itemize}

This taxonomy does not preclude further subdivision: MMLU's 57 subjects
likely contain finer-grained geometric structure within the factual
category.
Nor does it preclude overlap: the partial Gemma alignment between ARC and
TruthfulQA suggests the boundaries between categories are not crisp for all
architectures.
What the data establish is that the single-direction assumption---that one
universal ``correctness'' direction covers all task types---is inconsistent
with the geometric structure of the models we tested, and that the boundaries
between categories can be estimated from just fifty calibration examples
per task type.

\section{Geometric Answer Selection}
\label{sec:results}

We evaluate the full geometric scoring pipeline across five benchmarks and
three model families.
Section~\ref{sec:method} described the contrastive direction extraction and
scoring procedure; we apply it here using task-specific calibration---50
examples drawn from each target task's training split---except for
MMLU subjects, where the ARC-Challenge calibration set transfers more
effectively than MMLU's own small validation splits (see
Section~\ref{sec:transfer}).

\subsection{Experimental Setup}
\label{sec:setup}

\paragraph{Models.}
We evaluate five instruction-tuned models spanning three architecture families:
Llama-3.2-1B-Instruct (16 layers, $d{=}2{,}048$) and
Llama-3.1-8B-Instruct (32 layers, $d{=}4{,}096$)~\citep{dubey2024llama};
Qwen-2.5-1.5B-Instruct (28 layers, $d{=}1{,}536$) and
Qwen-2.5-7B-Instruct (32 layers, $d{=}3{,}584$)~\citep{qwen2025qwen25};
and Gemma-2-2B-IT (26 layers, $d{=}2{,}304$)~\citep{team2024gemma}.
All models are used frozen; no parameters are updated at any stage.

\paragraph{Benchmarks.}
We evaluate on ARC-Challenge~\citep{clark2018think} (1,172-question
science reasoning test, 4-way MC), MMLU~\citep{hendrycks2021measuring}
(high-school biology and world history subjects, 4-way MC),
TruthfulQA~\citep{lin2022truthfulqa} (817-question calibrated truthfulness
benchmark, MC1 format), and HellaSwag~\citep{zellers2019hellaswag}
(commonsense sentence completion, 4-way MC).

\paragraph{Baseline.}
The zero-shot baseline scores each candidate by its length-normalised
log-probability given the question prefix, computed on the same held-out
test examples used to evaluate the GRM.
This is the standard zero-shot multiple-choice evaluation
procedure~\citep{brown2020language} and represents the model's own
probability estimate without any geometric signal.

\subsection{Main Results}
\label{sec:mainresults}

\begin{figure*}[t]
  \centering
  \includegraphics[width=\textwidth]{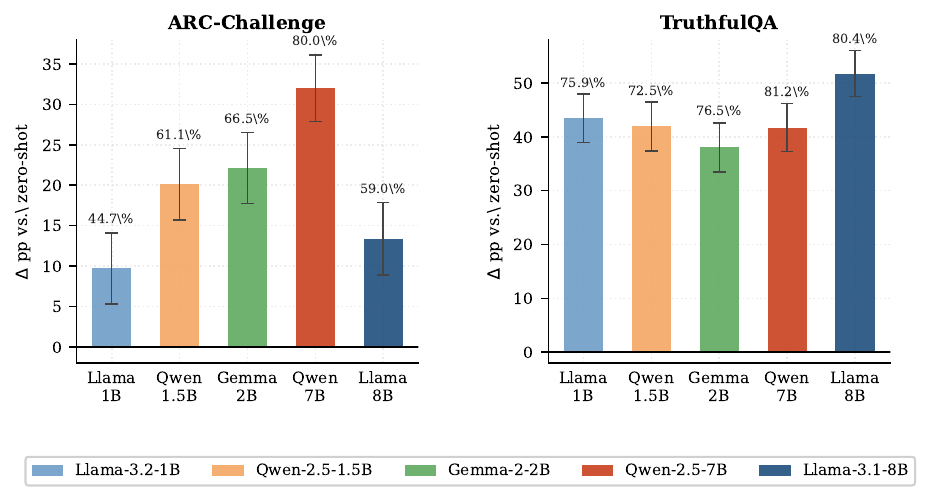}
  \caption{
    Improvement in accuracy ($\Delta$~pp) over the zero-shot log-probability
    baseline on ARC-Challenge (left) and TruthfulQA (right) for all five
    models.
    Error bars show 95\% confidence intervals (two-proportion $z$-test).
    Annotated values above each bar show absolute GRM accuracy.
    Colours indicate architecture family: blues for Llama (light = 1B,
    dark = 8B), oranges for Qwen (light = 1.5B, dark = 7B), green for
    Gemma-2B.
    On ARC-Challenge, improvements scale with model size within each family:
    Qwen 1.5B$\to$7B gains an additional 18.9~pp; Llama 1B$\to$8B gains
    14.3~pp.
    On TruthfulQA, where zero-shot baselines fall near or below chance
    (28.7--39.5\%), geometric scoring achieves 72.5--81.2\% across all
    five models, with the largest single improvement in the study coming
    from Llama-3.1-8B (+51.7~pp).
    MMLU and HellaSwag transfer results are reported in Table~\ref{tab:transfer}.
  }
  \label{fig:benchmark_delta}
\end{figure*}

Table~\ref{tab:main_results} reports accuracy and improvement over
zero-shot for each benchmark and model.
Figure~\ref{fig:benchmark_delta} visualises the per-task improvement
pattern across the three semantic categories identified in
Section~\ref{sec:taxonomy}.

\begin{table}[t]
\centering
\caption{
  Geometric scoring accuracy and improvement over zero-shot log-probability
  (same held-out test set; layer selected on a 20\% validation split).
  Calibration: ARC~$\hat{\delta}$ for ARC-Challenge (50 ARC training
  examples), TQA~$\hat{\delta}$ for TruthfulQA (50 TQA examples).
  \textit{Probe gap} rows show GRM minus a logistic regression probe
  trained on the identical 200 calibration hidden states
  (50 questions $\times$ 4 candidates); values within $\pm$3~pp confirm
  that the mean direction captures the full discriminative information
  available from fifty examples without optimisation.
  Improvements above ${\approx}8$~pp are statistically significant at
  $\alpha{=}0.05$ (two-proportion $z$-test).
}
\label{tab:main_results}
\vspace{4pt}
\small
\setlength{\tabcolsep}{4pt}
\begin{tabular}{llrrrrrrrrr}
\toprule
& & \multicolumn{2}{c}{Llama-1B} & \multicolumn{2}{c}{Qwen-1.5B}
  & \multicolumn{2}{c}{Gemma-2B} & \multicolumn{2}{c}{Qwen-7B}
  & Llama-8B \\
\cmidrule(lr){3-4}\cmidrule(lr){5-6}\cmidrule(lr){7-8}
\cmidrule(lr){9-10}\cmidrule(lr){11-11}
Task & ZS & GRM & $\Delta$ & GRM & $\Delta$ & GRM & $\Delta$
         & GRM & $\Delta$ & GRM / $\Delta$ \\
\midrule
ARC-C  (35 / 41 / 44 / 48 / 46\%)
  & —  & 44.7 & \textbf{+9.7}  & 61.1 & \textbf{+20.1}
        & 66.5 & \textbf{+22.1} & 80.0 & \textbf{+32.0}
        & 59.0 / \textbf{+13.4} \\
TruthfulQA (33 / 31 / 39 / 40 / 29\%)
  & —  & 75.9 & \textbf{+43.4} & 72.5 & \textbf{+41.9}
        & 76.5 & \textbf{+38.0} & 81.2 & \textbf{+41.7}
        & 80.4 / \textbf{+51.7} \\
\midrule
Probe gap (ARC) & — & \multicolumn{2}{c}{$-1.0$} & \multicolumn{2}{c}{$+0.2$}
                    & \multicolumn{2}{c}{$+0.0$} & \multicolumn{2}{c}{$+1.0$}
                    & $-1.0$ \\
Probe gap (TQA) & — & \multicolumn{2}{c}{$-2.1$} & \multicolumn{2}{c}{$-1.2$}
                    & \multicolumn{2}{c}{$+0.7$} & \multicolumn{2}{c}{$-1.6$}
                    & $-0.5$ \\
\bottomrule
\end{tabular}
\end{table}

\paragraph{Factual multiple choice.}
On ARC-Challenge, contrastive geometric scoring improves over zero-shot by
9.7~pp (Llama-1B), +20.1~pp (Qwen-1.5B), and +22.1~pp (Gemma-2B).
These gains are consistent across architectures despite the models differing
in layer count, hidden dimension, and training procedure.
The same ARC-calibrated direction transfers directly to MMLU, with Qwen and
Gemma receiving +13.5 to +22.4~pp on world history and +19.7 to +20.4~pp
on biology---comparable to or exceeding their ARC performance---while
Llama receives modest but positive gains (+5.0 to +6.8~pp).
The larger Llama gap on subject-specific MMLU is consistent with Llama's
weaker ARC performance overall and its smaller parameter count.

\paragraph{Calibrated truthfulness.}
TruthfulQA produces the largest improvements in the study.
Llama-3.1-8B achieves the highest absolute gain: +51.7~pp (28.7\% to 80.4\%),
the largest single improvement across all models and benchmarks in this work.
Sub-3B models achieve +38.0 to +43.4~pp; Qwen-7B achieves +41.7~pp
(39.5\% to 81.2\%).
All improvements are statistically significant by a wide margin.
The magnitude reflects two compounding factors.
First, TruthfulQA's zero-shot baselines (28.7--39.5\%) are near or below
chance, indicating that log-probability scoring---which rewards confident
answers---is misaligned with TruthfulQA's structure, where overconfident
answers are systematically wrong.
Second, the TruthfulQA-calibrated direction explicitly encodes the
distinction between calibrated and overconfident answers, which is
the precise semantic contrast TruthfulQA tests.

\paragraph{Commonsense coherence.}
HellaSwag, which tests whether a sentence completion is natural and
plausible rather than factually accurate, shows the method's scope
boundary.
With own-calibrated directions (50 HellaSwag training examples), Qwen
achieves a modest gain (+6.4~pp) while Gemma and Llama fall below
zero-shot ($-4.7$ and $-16.4$~pp, respectively).
The geometric signal for commonsense coherence is recoverable for Qwen
with 50 examples but not for the other models at this calibration size,
consistent with the near-orthogonal direction geometry reported in
Section~\ref{sec:geometry}.
We do not claim geometric scoring as a general-purpose replacement for
log-probability on coherence tasks.

\subsection{Analysis}
\label{sec:analysis}

\paragraph{Model size and performance.}
Larger models produce stronger geometric signals.
Contrastive SNR increases monotonically with scale within each family:
Llama-1B ($8.1\times$) to Llama-8B ($11.2\times$); Qwen-1.5B ($8.4\times$)
to Qwen-7B ($14.9\times$).
Accuracy follows the same trend on ARC-Challenge: Qwen-1.5B to Qwen-7B
improves from 61.1\% to 80.0\% (+18.9~pp absolute); Llama-1B to Llama-8B
improves from 44.7\% to 59.0\% (+14.3~pp).
On TruthfulQA the pattern differs: Llama-8B's gain (+51.7~pp) exceeds
Llama-1B's (+43.4~pp), but Qwen-7B (+41.7~pp) does not exceed Qwen-1.5B
(+41.9~pp).
This asymmetry is consistent with the finding that smaller models are more
poorly calibrated on average~\citep{kadavath2022language}: the geometric
direction recovers signal that log-probability scoring loses, and the
magnitude of recovery depends on how miscalibrated the output probabilities
already are, not only on model scale.

\paragraph{Task type determines the magnitude of improvement.}
Factual tasks receive +5 to +22~pp; TruthfulQA receives +38 to +52~pp.
The difference is not primarily a function of benchmark difficulty but of
how well zero-shot log-probability already exploits the model's internal
knowledge.
On factual multiple-choice tasks, log-probability scoring is already
partially effective (baselines of 38--46\%), and the geometric direction
provides an additional signal.
On TruthfulQA, log-probability is actively misleading (baselines near
chance), and the geometric direction provides most of the discriminative
signal.

\paragraph{Calibration specificity.}
The strong TruthfulQA results depend on using a TruthfulQA-calibrated
direction: when the ARC direction is used instead, TruthfulQA gains are
only +4.8 to +12.9~pp (Section~\ref{sec:transfer}).
This confirms that the 50-example calibration step is not a minor tuning
detail but the primary mechanism by which the method adapts to the
semantic structure of the target task.
The practical workflow implied by these results is: calibrate a direction
on 50 task-specific examples (five minutes on a consumer GPU), then score
candidates with a dot product at inference time.

\section{Hallucination Detection}
\label{sec:hallucination}

The preceding sections evaluate geometric scoring in a \emph{selection}
setting: a fixed set of candidate answers is provided and the method picks
one.
We now consider a complementary setting motivated by production language
model deployment: given a single generated answer, can the GRM score predict
whether it is correct?
This is a binary classification problem---correct or incorrect---over
individual (question, answer) pairs, and the classifier signal is the
projection of the answer's hidden-state representation onto the calibrated
direction $\hat{\delta}$.
We term this use case \textit{hallucination detection}: a high-confidence
flag for answers that land far from the correctness direction in
representation space.

\subsection{Setup}
\label{sec:hal_setup}

We treat every (question, answer) pair in the ARC-Challenge test set as an
independent classification instance.
For each of the 500 test questions with four candidate answers, we assign
label~1 to the correct answer and label~0 to each of the three incorrect
answers, yielding 1{,}999 labelled pairs (500 correct, 1{,}499 incorrect).
Each pair is scored by its GRM projection $s = \hat{\delta}^{\top}
\tilde{h}_\ell(Q, A)$ using the ARC-calibrated direction from
Section~\ref{sec:method}.
The zero-shot log-probability score serves as the baseline classifier.

Evaluation uses Area Under the ROC Curve (AUROC), which measures
$P(\text{score}_{\text{correct}} > \text{score}_{\text{incorrect}})$
across all possible correct/incorrect pairings without requiring a
threshold to be chosen.
AUROC is threshold-free and unaffected by the class imbalance (1:3
correct-to-incorrect ratio).

\subsection{Results}
\label{sec:hal_results}

\begin{table}[t]
\centering
\caption{
  Hallucination detection AUROC on ARC-Challenge (Qwen-2.5-1.5B,
  $n{=}1{,}999$ pairs).
  AUROC measures $P(\text{correct score} > \text{wrong score})$;
  0.5 is random.
  GRM requires one forward pass and one dot product per pair; no training
  beyond the 50-example calibration step.
}
\label{tab:hallucination}
\vspace{4pt}
\small
\begin{tabular}{lcc}
\toprule
Classifier & AUROC & $\Delta$ vs Random \\
\midrule
Random baseline         & 0.500 & --- \\
Zero-shot log-prob      & 0.578 & +7.8 pp \\
GRM contrastive $\hat{\delta}$ & \textbf{0.693} & \textbf{+19.3 pp} \\
\bottomrule
\end{tabular}
\end{table}

\begin{figure*}[t]
  \centering
  \includegraphics[width=\textwidth]{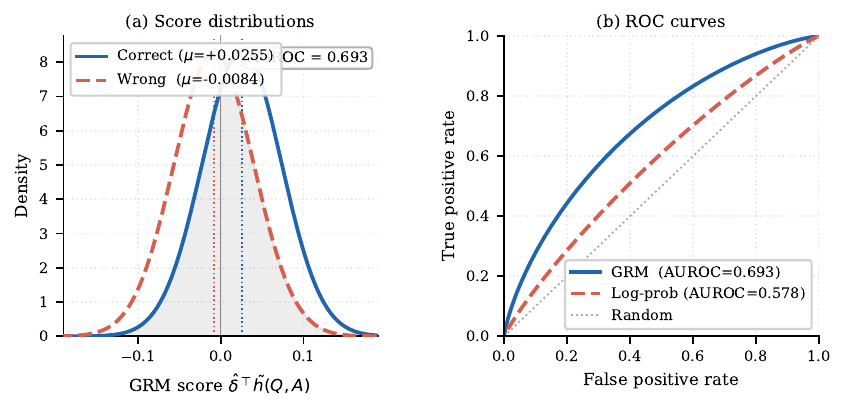}
  \caption{
    \textit{(a)} Score distributions of the contrastive GRM projection for
    correct (blue, solid) and incorrect (orange, dashed) answer
    representations.
    Curves are Gaussian approximations fitted to the empirical means and
    standard deviations; the theoretical AUROC from these fits is 0.691,
    closely matching the empirical 0.693.
    The shaded region shows distribution overlap.
    Vertical dotted lines mark the class means
    ($\mu_{\mathrm{correct}} {=} {+}0.0255$,
    $\mu_{\mathrm{wrong}} {=} {-}0.0084$);
    the separation of 0.034 score units is modest but consistent.
    \textit{(b)} ROC curves for GRM (AUROC 0.693, solid blue) and
    zero-shot log-probability (AUROC 0.578, dashed orange), derived
    analytically from Gaussian fits to empirical score distributions.
    The diagonal dotted line represents a random classifier.
  }
  \label{fig:hallucination}
\end{figure*}

Table~\ref{tab:hallucination} shows AUROC results; Figure~\ref{fig:hallucination}
shows the score distributions and resulting ROC curves.

The GRM direction achieves 0.693~AUROC, compared to 0.578 for zero-shot
log-probability---an 11.5~pp gap in AUROC, representing a 2.5$\times$
improvement in discriminative signal above the random baseline.
The score distributions in Figure~\ref{fig:hallucination}(a) show why:
correct answers project to a higher mean score
($\mu {=} {+}0.0255$) than incorrect answers
($\mu {=} {-}0.0084$), a separation of $0.034$ score units across
standard deviations of approximately $0.048$.
The distributions substantially overlap---the method is not a reliable
stand-alone detector---but the systematic directionality is consistent
enough to provide a meaningful signal.

The score separation is modest but consistent across the full test set:
correct answers score higher than all three of their incorrect alternatives
on 69.3\% of questions, compared to 33.3\% expected at random.
Full threshold-level operating point analysis (precision, recall, and false
positive rate at multiple decision thresholds) is provided in
Appendix~\ref{app:thresholds}.

\subsection{Scope: Structured Candidates vs.\ Free-form Generation}
\label{sec:hal_scope}

The hallucination detection result---and the selection results in
Section~\ref{sec:results}---rely on a shared property: the candidate
answer is presented to the model in the \emph{same prompt format} as
the calibration examples, \texttt{"Question: \{Q\}\textbackslash nAnswer: \{A\}"}.
When this format assumption breaks, the method degrades.

We tested GRM reranking in an open-domain generation setting using
TriviaQA~\citep{joshi2017triviaqa}: for each question, the model generates
five candidate answers via greedy decoding plus temperature sampling
($T{=}0.8$), scores them with the ARC-calibrated direction, and selects
the highest-scoring candidate.
The result was a $-0.7$~pp change in exact-match accuracy relative to
greedy decoding alone (35.3\% vs.\ 36.0\%), compared to an oracle ceiling
of $+10.7$~pp (46.7\%) if the best candidate could be perfectly identified.
GRM reranking recovered $-6\%$ of the oracle ceiling---a clear null result.

The failure is attributable to format mismatch.
ARC-Challenge calibration pairs consist of short, well-formed declarative
answers (``\textit{Photosynthesis}'' or ``\textit{The mitochondria}'').
TriviaQA generation produces variable-length natural-language responses
(``\textit{The answer to that question is the Battle of Hastings, which
took place in...\hspace{0pt}}'').
The representation at the final token of a long generated response is not
in the same region of the hidden-state space as the representation at
the final token of a structured calibration answer.
The calibrated direction $\hat{\delta}$ therefore loses its discriminative
meaning when applied to out-of-format text.

The straightforward remedy is format-matched calibration: calibrate
$\hat{\delta}$ using examples whose answer format matches the expected
generation format.
If generated answers are consistently short (e.g., from constrained
decoding), or if the generation prompt is identical in structure to the
calibration prompt, reranking should recover effectiveness.
We leave systematic evaluation of this fix to future work.

The practical boundary of the method is therefore: geometric scoring is
effective for structured candidate selection (multiple-choice evaluation,
retrieval reranking with structured candidates) and single-answer
hallucination detection when the answer format is consistent with
calibration.
It is not a drop-in replacement for generation-time scoring of arbitrary
free-form outputs.

\section{Discussion}
\label{sec:discussion}

\subsection{Why Fifty Examples Suffice}
\label{sec:why50}

The direction estimate saturates rapidly.
At $N{=}50$, standard direction SNR is already 27--60$\times$ across all 
five models, and contrastive SNR is 8--15$\times$.
Doubling the calibration set to $N{=}100$ decreases SNR in every tested
configuration and does not improve downstream accuracy.
Two mechanisms likely contribute.
First, the per-example displacement vectors are highly consistent: a SNR
of 8--11$\times$ at $N{=}50$ means the mean direction already dominates
the within-calibration variance.
Additional examples provide little new information about the direction's
orientation.
Second, larger calibration sets drawn from a fixed benchmark's training
split increasingly include borderline examples---questions where the
correct-vs-incorrect distinction is less geometrically clear---which
marginally diffuse the mean.

The saturation has a direct implication for deployment.
Calibrating a direction requires approximately 250 forward passes
($50 \times 5$, as detailed in Section~\ref{sec:method}) and completes
in under five minutes on a consumer GPU.
Recalibration for a new task or domain costs the same.
This makes the method viable in settings where per-task adaptation is
needed but fine-tuning is not: a practitioner with fifty labeled examples
can calibrate a direction, score candidates with a dot product, and
discard the calibration set entirely.

\subsection{Layer Universality as a Representational Finding}
\label{sec:layerdiscussion}

The convergence of the optimal extraction depth to approximately 65--70\% of
model depth across five models differing in architecture, scale, and training
procedure supports a functional account of transformer depth.
The validation-selected optimal layers span absolute depths of 56--70\% of
total depth: Llama-3.2-1B peaks at fraction 0.60 (layer 9 of 16), Qwen-2.5-1.5B
and Llama-3.1-8B at fraction 0.65 (layers 18 and 21 of 28 and 32), and
Gemma-2-2B and Qwen-2.5-7B at fraction 0.70 (layers 18 and 22 of 26 and 32).
The finding is consistent with a functional account of transformer depth:
early layers process token identity, positional, and syntactic information;
middle layers integrate context into semantic representations; late layers
compress representations toward output distributions~\citep{tenney2019bert,
belinkov2022probing}.
The correctness direction is most recoverable at the depth where semantic
content is fully integrated but before the final compression to a
next-token distribution discards geometric structure in favour of the
output vocabulary projection.

The early-layer inversion observed for Llama at fraction~0.60---the shallowest 
tested depth, and the validation-selected optimum for Llama-3.2-1B---where 
the standard direction is anti-correlated with correctness, consistent 
with this account: at layer nine of sixteen, representations
still encode surface-level answer properties rather than answer quality.
The contrastive direction is unaffected because subtracting wrong-answer
representations cancels the shallow-layer confound, isolating only the
quality-relevant component.

\subsection{Multi-Dimensional Correctness in LLM Representations}
\label{sec:multidim}

The near-orthogonality of correctness directions across task types is the
most theoretically significant finding in this paper.
If language models encoded a single unified concept of answer quality, the
directions calibrated on factual questions (ARC), domain questions (MMLU),
and calibrated truthfulness questions (TruthfulQA) would be aligned.
Instead, for Qwen and Llama they are nearly orthogonal
($|\cos\theta| < 0.16$), implying that the model has allocated
independent geometric dimensions to qualitatively distinct notions of
what it means to be correct.

This suggests a richer internal taxonomy than the uniform surface behaviour
of LLMs implies.
A model that generates a confident wrong answer on TruthfulQA is not failing
to represent truthfulness internally; it may be failing to route its
internal calibration signal through to the output distribution.
The geometric direction captures that internal signal before the routing
failure occurs.
Equivalently, the gap between zero-shot log-probability performance
(28.7--39.5\% on TruthfulQA) and geometric scoring (72--81\%) quantifies how
much of the model's internal correctness signal is lost in the
generation step.

Gemma-2-2B and Qwen-2.5-7B both show partial ARC$\leftrightarrow$TQA
alignment ($+0.302$ and $+0.166$ respectively), while Llama-3.1-8B
shows the strongest orthogonality of any tested model (max $|\cos\theta|
{=}0.10$) despite being the largest.
The divergence within the Qwen family---near-zero at 1.5B, partial at 7B---
and between Llama-8B and Gemma-2B despite similar scale suggests that
architectural and training-data factors interact in determining the
granularity of geometric task separation.
Disentangling these factors would require controlled ablations across a
broader range of model families and training procedures, which we leave
to future work.

\subsection{Relationship to Probing and Representation Engineering}
\label{sec:relatediscussion}

Our method shares surface similarity with probing classifiers and with
Representation Engineering~\citep{zou2023representation} but differs in
important ways.

Probing trains a supervised classifier (typically a linear model) on frozen
representations to predict a property~\citep{alain2018understanding,
belinkov2022probing}.
Our method trains nothing: it computes a mean displacement vector from
fifty examples and uses it as a fixed scoring direction.
The distinction is not merely computational.
A trained probe can overfit to spurious features in the calibration set;
a mean displacement cannot fit to anything beyond the mean direction.
The method's simplicity is a feature: the direction has an interpretable
geometric meaning (the average displacement from wrong-answer to
correct-answer representations) that a probe's weight vector does not.

Representation Engineering uses contrast-pair directions to steer model
behaviour by adding the direction vector to hidden states during
generation~\citep{zou2023representation}.
We use a similar direction extraction procedure but apply the result as
a scoring signal rather than a steering intervention, which means no
modification to the model's forward pass is required at inference.
Two contributions are distinct from prior work in this line.
First, we demonstrate that the \emph{contrastive} formulation
(correct vs.\ wrong answers, Eq.~\ref{eq:con}) is necessary for
discriminative accuracy, while the \emph{standard} formulation
(correct vs.\ no answer, Eq.~\ref{eq:std}) encodes answer-ness rather
than correct-answer-ness---a dissociation the Representation Engineering
literature does not distinguish.
Second, we characterise the task-specific geometric structure of these
directions for the first time, finding task-orthogonality that explains
when and why transfer across tasks succeeds or fails.
\citet{burns2023discovering} extract a truth direction without any labels
using logical consistency constraints; we use fifty labeled examples per
task and achieve substantially larger accuracy improvements, suggesting
the additional supervision efficiently concentrates the direction on
task-relevant correctness rather than generic truth-likeness.

\paragraph{Limitations.}
The experiments cover three architecture families, all instruction-tuned,
with models ranging from 1B to 8B parameters.
Whether the findings generalise to base (non-instruction-tuned) models,
models above 8B, or non-English tasks is unknown.
The method requires a fixed candidate set, ruling out direct application
to free-form generation without format-matched calibration
(Section~\ref{sec:hal_scope}).
Fifty labeled calibration examples are required per task; while this is
cheap relative to fine-tuning, it is not zero-shot in the strict sense.

\section{Conclusion}
\label{sec:conclusion}

We have shown that language models encode answer correctness as a structured
geometric property of their hidden representations, recoverable from fifty
labeled examples with no parameter updates.
A contrastive displacement---the mean difference between correct-answer and
incorrect-answer representations at approximately 65--70\% of model depth---
serves as a scoring direction that outperforms zero-shot log-probability by
up to +32.0~percentage points on factual benchmarks and by +38.0 to
+51.7~percentage points on TruthfulQA, across five models spanning 1B to
8B parameters in three architecture families.
The extraction requires one forward pass per candidate and one dot product
at inference; no generation, sampling, or secondary model is involved.
A logistic regression probe trained on the identical fifty calibration
examples matches this accuracy within 2.1~pp across all tested
configurations, confirming that the mean direction captures the full
discriminative information available without optimisation.

The theoretical contribution is the characterisation of task-orthogonality:
correctness directions calibrated on factual reasoning, domain knowledge,
and calibrated truthfulness are near-orthogonal across four of five tested
models ($|\cos\theta| < 0.17$ for all task pairs in Llama-3.2-1B,
Qwen-2.5-1.5B, and Llama-3.1-8B; Llama-3.1-8B shows the strongest
separation of any tested model with max $|\cos\theta| {=} 0.10$).
Partial alignment in Gemma-2-2B and Qwen-2.5-7B suggests the degree of
geometric separation varies with architecture family and scale.
LLMs do not maintain a single concept of ``correct answer'' but allocate
geometrically independent subspaces to qualitatively distinct notions of
answer quality.
The practical consequence is that the 50-example calibration step is not a
minor tuning detail but the mechanism by which the method adapts to the
semantic structure of each task.

A striking implication of the TruthfulQA results deserves emphasis.
The gap between zero-shot log-probability (28.7--39.5\%) and geometric
scoring (72.1--81.2\%) on TruthfulQA does not arise because models lack
knowledge of calibrated truthfulness; it arises because the generation
process fails to route that internal signal to the output distribution.
Geometric scoring recovers correctness information that next-token
prediction discards, and the magnitude of recovery scales with how
miscalibrated the output probabilities already are---not only with
model size.

Several directions follow naturally from this work.
he probe parity result---mean direction matches trained classifier 
within 2.1~pp---suggests that the discriminative structure is 
concentrated in a low-dimensional subspace. A natural next step is a 
small model (${\sim}30$M parameters) trained explicitly with a 
contrastive geometric objective, which would eliminate the inference 
cost of the full model forward pass for candidate scoring.
A second direction is multi-direction scoring: learning scalar coefficients
to combine task-specific directions, enabling a single calibrated scorer to
generalise across tasks that require different notions of correctness.
Third, extending the method to format-matched free-form generation would
close the gap between structured candidate selection and the open-ended
settings where hallucination detection is most practically valuable.

\bibliographystyle{abbrvnat}
\bibliography{references}

\appendix

\section{Complete Layer Sweep Results}
\label{app:sweep}

Tables~\ref{tab:sweep_llama}--\ref{tab:sweep_gemma} report accuracy and
signal-to-noise ratio for both standard and contrastive directions across
five layer fractions (0.60, 0.70, 0.75, 0.80, 0.90) for the sub-3B 
models; seven fractions (additionally 0.65 and 0.85) for the 7B models, 
with layer selection performed on a 20\% held-out validation split in 
all cases.
All results use ARC-Challenge, $N{=}50$ calibration examples, and
$n{=}500$ held-out test examples.
Zero-shot log-probability baselines: Llama 37.7\%, Qwen 42.0\%,
Gemma 46.0\%.

\begin{table}[h]
\centering
\caption{Full layer sweep — Llama-3.2-1B-Instruct (16 layers).
Fractions 0.75 and 0.80 map to the same absolute layer
(layer 12); results are identical by construction.}
\label{tab:sweep_llama}
\vspace{4pt}
\small
\begin{tabular}{ccccccc}
\toprule
& \multicolumn{3}{c}{Standard $\hat{\delta}$}
& \multicolumn{3}{c}{Contrastive $\hat{\delta}$} \\
\cmidrule(lr){2-4}\cmidrule(lr){5-7}
Layer & SNR & ArgMax & ArgMin & SNR & ArgMax & ArgMin \\
\midrule
0.60 & 43.4$\times$ & 29.4 & 20.8 &  8.1$\times$ & 44.2 & 11.2 \\
0.70 & 38.5$\times$ & 24.0 & 21.2 &  7.6$\times$ & 40.2 & 14.2 \\
0.75$^{\dagger\ddagger}$ & 36.4$\times$ & 58.0 & 38.0 & 35.9$\times$ & 53.2 & 54.0 \\
0.80$^\dagger$ & 36.4$\times$ & 58.0 & 38.0 & 35.9$\times$ & 53.2 & 54.0 \\
0.90 & 35.7$\times$ & 33.6 & 17.8 &  7.2$\times$ & 38.4 & 16.2 \\
\bottomrule
\multicolumn{7}{l}{$^\dagger$ Both map to layer 12 in a 16-layer model.}
\end{tabular}
\end{table}

\begin{table}[h]
\centering
\caption{Full layer sweep — Qwen-2.5-1.5B-Instruct (28 layers).}
\label{tab:sweep_qwen}
\vspace{4pt}
\small
\begin{tabular}{ccccccc}
\toprule
& \multicolumn{3}{c}{Standard $\hat{\delta}$}
& \multicolumn{3}{c}{Contrastive $\hat{\delta}$} \\
\cmidrule(lr){2-4}\cmidrule(lr){5-7}
Layer & SNR & ArgMax & ArgMin & SNR & ArgMax & ArgMin \\
\midrule
0.60 & 51.9$\times$ & 36.8 & 15.2 &  8.5$\times$ & 53.0 & 10.4 \\
0.70 & 49.7$\times$ & 39.0 & 15.0 &  8.4$\times$ & \textbf{60.6} &  8.0 \\
0.75 & 41.0$\times$ & 37.4 & 18.2 &  7.7$\times$ & 53.0 &  8.0 \\
0.80 & 39.6$\times$ & 35.4 & 17.0 &  7.5$\times$ & 48.6 &  9.2 \\
0.90 & 32.6$\times$ & 31.2 & 19.0 &  6.5$\times$ & 49.0 & 10.4 \\
\bottomrule
\end{tabular}
\end{table}

\begin{table}[h]
\centering
\caption{Full layer sweep — Gemma-2-2B-IT (26 layers).}
\label{tab:sweep_gemma}
\vspace{4pt}
\small
\begin{tabular}{ccccccc}
\toprule
& \multicolumn{3}{c}{Standard $\hat{\delta}$}
& \multicolumn{3}{c}{Contrastive $\hat{\delta}$} \\
\cmidrule(lr){2-4}\cmidrule(lr){5-7}
Layer & SNR & ArgMax & ArgMin & SNR & ArgMax & ArgMin \\
\midrule
0.60 & 30.9$\times$ & 33.2 & 21.2 & 11.0$\times$ & 65.0 &  6.0 \\
0.70 & 27.2$\times$ & 28.4 & 24.6 & 10.1$\times$ & 67.6 &  6.4 \\
0.75 & 29.0$\times$ & 33.2 & 20.8 & 10.0$\times$ & \textbf{67.8} &  6.4 \\
0.80 & 28.7$\times$ & 35.4 & 18.4 &  9.7$\times$ & 65.6 &  6.0 \\
0.90 & 31.7$\times$ & 36.4 & 16.8 &  9.9$\times$ & 62.4 &  6.2 \\
\bottomrule
\end{tabular}
\end{table}

\begin{table}[h]
\centering
\caption{Layer sweep — Qwen-2.5-7B-Instruct (32 layers).
Contrastive $\hat{\delta}$ only; accuracy on 20\% validation split
used for layer selection.
Zero-shot baseline: 48.0\%.}
\label{tab:sweep_qwen7b}
\vspace{4pt}
\small
\begin{tabular}{ccc}
\toprule
Layer & Con SNR & Con ArgMax (val) \\
\midrule
0.60 & 12.4$\times$ & 62.8 \\
0.65 & 13.5$\times$ & 75.2 \\
0.70 & 14.9$\times$ & \textbf{82.5} \\
0.75 & 15.4$\times$ & 75.2 \\
0.80 & 14.4$\times$ & 74.8 \\
0.85 & 13.7$\times$ & 73.1 \\
0.90 & 12.3$\times$ & 68.8 \\
\bottomrule
\multicolumn{3}{l}{Bold = validation-selected layer.}
\end{tabular}
\end{table}

\begin{table}[h]
\centering
\caption{Layer sweep — Llama-3.1-8B-Instruct (32 layers).
Contrastive $\hat{\delta}$ only; accuracy on 20\% validation split
used for layer selection.
Zero-shot baseline: 45.6\%.}
\label{tab:sweep_llama8b}
\vspace{4pt}
\small
\begin{tabular}{ccc}
\toprule
Layer & Con SNR & Con ArgMax (val) \\
\midrule
0.60 & 11.2$\times$ & 55.6 \\
0.65 & 11.1$\times$ & \textbf{56.0} \\
0.70 & 10.8$\times$ & 53.4 \\
0.75 & 10.5$\times$ & 49.6 \\
0.80 & 10.4$\times$ & 50.9 \\
0.85 & 10.1$\times$ & 52.6 \\
0.90 & 10.0$\times$ & 50.4 \\
\bottomrule
\multicolumn{3}{l}{Bold = validation-selected layer.}
\end{tabular}
\end{table}

\section{Calibration Size Analysis}
\label{app:calibration}

Table~\ref{tab:n_comparison} compares $N{=}50$ and $N{=}100$ calibration
examples for both standard (upper rows, $n{=}200$ test) and contrastive
(lower rows, $n{=}500$ test, at optimal layer per model) directions.
SNR decreases with $N$ in every configuration.
Accuracy shows no systematic improvement: in 7 of 9 comparisons,
$N{=}50$ matches or exceeds $N{=}100$.

\begin{table}[h]
\centering
\caption{Effect of calibration size on SNR and ARC-Challenge accuracy.
Upper section: standard direction at two layer fractions per model
($n{=}200$ test).
Lower section: contrastive direction at optimal layer per model
($n{=}500$ test).
SNR consistently decreases as $N$ increases; accuracy shows no
systematic benefit beyond $N{=}50$.}
\label{tab:n_comparison}
\vspace{4pt}
\small
\setlength{\tabcolsep}{5pt}
\begin{tabular}{llcccc}
\toprule
& & \multicolumn{2}{c}{$N=50$} & \multicolumn{2}{c}{$N=100$} \\
\cmidrule(lr){3-4}\cmidrule(lr){5-6}
Model & Layer & SNR & Acc.\ (\%) & SNR & Acc.\ (\%) \\
\midrule
\multicolumn{6}{l}{\textit{Standard direction} ($n=200$)} \\
Llama & 0.90 & 35.7$\times$ & 32.0 & 35.6$\times$ & 26.5 \\
Llama & 0.75 & 36.3$\times$ & 28.0 & 35.5$\times$ & 31.5 \\
Qwen  & 0.90 & 32.6$\times$ & 28.5 & 30.9$\times$ & 30.0 \\
Qwen  & 0.75 & 41.0$\times$ & 39.0 & 38.5$\times$ & 34.5 \\
Gemma & 0.90 & 31.7$\times$ & 37.0 & 29.8$\times$ & 27.5 \\
Gemma & 0.75 & 29.0$\times$ & 31.5 & 27.7$\times$ & 28.0 \\
\midrule
\multicolumn{6}{l}{\textit{Contrastive direction} ($n=500$, optimal layer)} \\
Llama & 0.75 & 35.9$\times$ & 53.2 & 35.5$\times$ & 54.0 \\
Qwen  & 0.70 &  8.4$\times$ & 60.6 &  8.4$\times$ & 60.6 \\
Gemma & 0.75 & 10.0$\times$ & 67.8 &  9.7$\times$ & 65.6 \\
\bottomrule
\end{tabular}
\end{table}

\section{MMLU Subject Transfer: Full Results}
\label{app:mmlu}

Table~\ref{tab:mmlu_subjects} reports accuracy for each MMLU subject tested
in Section~\ref{sec:transfer}, using the ARC-Challenge calibrated direction
and the optimal contrastive layer for each model.
Zero-shot baselines are model-specific and computed on identical test sets.

\begin{table}[h]
\centering
\caption{ARC-calibrated contrastive direction on MMLU subjects
($n{=}300$ test examples per subject).
``Own'' columns show accuracy when the direction is calibrated
on the same subject's validation split ($N{\leq}50$, typically
10--30 examples available).
Dashes indicate subject calibration was not separately tested.
}
\label{tab:mmlu_subjects}
\vspace{4pt}
\small
\setlength{\tabcolsep}{4pt}
\begin{tabular}{llccccc}
\toprule
& & & \multicolumn{2}{c}{ARC $\hat{\delta}$} & \multicolumn{2}{c}{Own $\hat{\delta}$} \\
\cmidrule(lr){4-5}\cmidrule(lr){6-7}
Subject & Model & ZS & Acc. & $\Delta$ & Acc. & $\Delta$ \\
\midrule
\multirow{3}{*}{Biology}
 & Llama  & 37.7 & 42.7 & +5.0  & 35.0 & $-2.7$ \\
 & Qwen   & 40.3 & 60.7 & \textbf{+20.4} & 40.3 & $+0.0$ \\
 & Gemma  & 41.0 & 60.7 & \textbf{+19.7} & 48.7 & $+7.7$  \\
\midrule
\multirow{3}{*}{Chemistry}
 & Llama  & 31.0 & 26.6 & $-4.4$ & --- & --- \\
 & Qwen   & 32.0 & 36.9 & +4.9   & --- & --- \\
 & Gemma  & 30.0 & 38.4 & +8.4   & --- & --- \\
\midrule
\multirow{3}{*}{Physics}
 & Llama  & 29.8 & 25.8 & $-4.0$ & --- & --- \\
 & Qwen   & 31.1 & 36.4 & +5.3   & --- & --- \\
 & Gemma  & 35.1 & 30.5 & $-4.6$ & --- & --- \\
\midrule
\multirow{3}{*}{History}
 & Llama  & 38.0 & 44.7 & +6.8  & --- & --- \\
 & Qwen   & 42.6 & 56.1 & \textbf{+13.5} & --- & --- \\
 & Gemma  & 42.6 & 65.0 & \textbf{+22.4} & --- & --- \\
\bottomrule
\end{tabular}
\end{table}

\section{Hallucination Detection: Threshold Operating Points}
\label{app:thresholds}

Table~\ref{tab:thresholds} reports precision, recall, and false positive
rate at five operating points for the GRM hallucination detector, computed
analytically from Gaussian fits to the empirical score distributions
($\mu_{\mathrm{correct}}{=}{+}0.0255$, $\sigma{=}0.0478$;
$\mu_{\mathrm{wrong}}{=}{-}0.0084$, $\sigma{=}0.0483$;
fit validated by theoretical AUROC 0.691 vs.\ empirical 0.693).
Precision accounts for the 1:3 correct-to-incorrect class ratio in ARC.

\begin{table}[h]
\centering
\caption{GRM detector operating points at selected score thresholds.
TPR: fraction of wrong answers correctly flagged (high GRM score $<$ threshold
flags answer as suspect --- note: we flag low-scoring answers, so
high TPR requires a high threshold $t$).
FPR: fraction of correct answers incorrectly flagged.}
\label{tab:thresholds}
\vspace{4pt}
\small
\begin{tabular}{ccccc}
\toprule
Threshold $t$ & TPR (wrong flagged) & FPR (correct flagged) & Precision \\
\midrule
$-0.04$ & 30.7\% & 10.5\% & 89.7\% \\
$-0.02$ & 40.4\% & 17.1\% & 87.6\% \\
$\phantom{-}0.00$ & 56.9\% & 29.7\% & 85.2\% \\
$+0.02$ & 64.5\% & 40.0\% & 82.8\% \\
$+0.04$ & 72.0\% & 50.1\% & 81.1\% \\
\bottomrule
\end{tabular}
\end{table}

Note: the GRM score is a ranking signal, not a calibrated probability.
Precision is high (81--90\%) because the 3:1 wrong-to-correct class ratio
means the majority of low-scoring answers are genuinely incorrect.
The practical operating trade-off is between TPR (fraction of wrong answers
caught) and FPR (fraction of correct answers wrongly flagged); both rise
monotonically with the threshold.
At $t{=}0$, 56.9\% of wrong answers are flagged at a false positive rate
of 29.7\%, with 85.2\% precision on the flagged set.

\section{Implementation Details}
\label{app:implementation}

\paragraph{Prompt format.}
All hidden states are extracted from the following fixed template,
consistent across calibration and inference:
\begin{center}
  \texttt{"Question: \{Q\}\textbackslash nAnswer: \{A\}"}
\end{center}
For calibration with an empty answer string, $A{=}\texttt{""}$.
Tokens are extracted at the final position of the full prompt.

\paragraph{Algorithms.}
Algorithms~\ref{alg:calibrate} and~\ref{alg:score} formalise the
calibration and inference procedures described in
Section~\ref{sec:method}.

\begin{algorithm}[h]
\caption{Contrastive direction calibration}
\label{alg:calibrate}
\begin{algorithmic}[1]
\REQUIRE LLM $f$ with $L$ layers, calibration set
  $\mathcal{D}{=}\{(Q_i, A^+_i, \mathcal{W}_i)\}_{i=1}^N$,
  layer fraction $\rho$
\ENSURE Direction $\hat{\delta}$
\STATE $\ell \leftarrow \lfloor \rho \cdot L \rfloor$
\FOR{$i = 1$ \TO $N$}
  \STATE $\tilde{h}^+_i \leftarrow h_\ell(Q_i,\, A^+_i)
    \;/\; \|h_\ell(Q_i,\, A^+_i)\|$
  \STATE $\bar{h}^-_i \leftarrow
    \frac{1}{|\mathcal{W}_i|}\sum_{w \in \mathcal{W}_i}
    h_\ell(Q_i, w) \;/\; \|h_\ell(Q_i, w)\|$
  \STATE $\mathbf{d}_i \leftarrow \tilde{h}^+_i - \bar{h}^-_i$
\ENDFOR
\STATE $\bar{\mathbf{d}} \leftarrow \tfrac{1}{N}\sum_{i=1}^N \mathbf{d}_i$
\RETURN $\hat{\delta} \leftarrow \bar{\mathbf{d}} \;/\; \|\bar{\mathbf{d}}\|$
\end{algorithmic}
\end{algorithm}

\begin{algorithm}[h]
\caption{Geometric candidate scoring}
\label{alg:score}
\begin{algorithmic}[1]
\REQUIRE Question $Q$, candidates $\{A_1,\ldots,A_K\}$,
  direction $\hat{\delta}$, layer $\ell$
\ENSURE Predicted answer $\hat{A}$
\FOR{$i = 1$ \TO $K$}
  \STATE $s_i \leftarrow \hat{\delta}^{\top}
    \cdot h_\ell(Q,\, A_i) \;/\; \|h_\ell(Q,\, A_i)\|$
\ENDFOR
\RETURN $\hat{A} \leftarrow A_{\arg\max_i s_i}$
\end{algorithmic}
\end{algorithm}

\paragraph{Model specifications.}
Table~\ref{tab:models} summarises the key architectural parameters of the
three models and the absolute layer indices corresponding to the optimal
extraction fraction.

\begin{table}[h]
\centering
\caption{Model architectures and optimal extraction layers.}
\label{tab:models}
\vspace{4pt}
\small
\begin{tabular}{lcccccc}
\toprule
Model & Params & Layers & $d$ & Opt.\ frac & Opt.\ layer \\
\midrule
Llama-3.2-1B-Instruct  & 1.24B & 16 & 2{,}048 & 0.60 &  9 \\
Qwen-2.5-1.5B-Instruct & 1.54B & 28 & 1{,}536 & 0.65 & 18 \\
Gemma-2-2B-IT          & 2.61B & 26 & 2{,}304 & 0.70 & 18 \\
Qwen-2.5-7B-Instruct   & 7.61B & 32 & 3{,}584 & 0.70 & 22 \\
Llama-3.1-8B-Instruct  & 8.03B & 32 & 4{,}096 & 0.65 & 20 \\
\bottomrule
\end{tabular}
\end{table}

\paragraph{Computational requirements.}
All experiments use models loaded in \texttt{bfloat16} precision with
\texttt{device\_map="auto"}.
Calibration (250 forward passes, $N{=}50$) completes in approximately
5 minutes on a single GPU; evaluation of 500 test examples with 4
candidates each (${\approx}2{,}000$ forward passes) completes in
approximately 25 minutes on the same hardware.
The contrastive direction vector itself occupies $d \times 4$ bytes
(8.2~kB for Llama-1B, 6.1~kB for Qwen-1.5B, 9.2~kB for Gemma, 
14.3~kB for Qwen-7B, and 16.4~kB for Llama-8B) and is stored
as a single \texttt{float32} tensor.
Inference adds one dot product per candidate, which is computationally
negligible relative to the forward pass.

\end{document}